\documentclass{ecai}

\usepackage{latexsym}
\usepackage{amssymb}
\usepackage{amsmath}
\usepackage{amsthm}
\usepackage{booktabs}
\usepackage{enumitem}
\usepackage{graphicx}
\usepackage{color}

\usepackage{float}

\usepackage{algorithm,algorithmic}
\makeatletter
\newcommand{\ALOOP}[1]{\ALC@it\algorithmicloop\ #1%
  \begin{ALC@loop}}
\newcommand{\ENDALOOP}{\end{ALC@loop}\ALC@it\algorithmicendloop}

\makeatother

\newcommand{\bg}[1]{\boldsymbol{#1}} 
\newcommand{\bm}[1]{\mathbf{#1}} 

\newcommand\T{{\mathpalette\raiseT\intercal}}
\newcommand\raiseT[2]{%
\setbox0\hbox{$#1{#2}$}\raise\dp0\box0}

\usepackage{booktabs} 
\usepackage{multirow}

\newcommand{\BibTeX}{B\kern-.05em{\sc i\kern-.025em b}\kern-.08em\TeX}

\begin{document}


\begin{frontmatter}


\paperid{1581}


\title{A Federated Large Language Model for\\ Long-Term Time Series Forecasting}

\author[1]{\fnms{Raed}~\snm{Abdel-Sater}}
\author[1]{\fnms{A.}~\snm{Ben Hamza}\thanks{Corresponding Author. Email: hamza@ciise.concordia.ca}}

\address[1]{Concordia University, Montreal, Canada}


\begin{abstract}
Long-term time series forecasting in centralized environments poses unique challenges regarding data privacy, communication overhead, and scalability. To address these challenges, we propose FedTime, a federated large language model (LLM) tailored for long-range time series prediction. Specifically, we introduce a federated pre-trained LLM with fine-tuning and alignment strategies. Prior to the learning process, we employ K-means clustering to partition edge devices or clients into distinct clusters, thereby facilitating more focused model training. We also incorporate channel independence and patching to better preserve local semantic information, ensuring that important contextual details are retained while minimizing the risk of information loss. We demonstrate the effectiveness of our FedTime model through extensive experiments on various real-world forecasting benchmarks, showcasing substantial improvements over recent approaches. In addition, we demonstrate the efficiency of FedTime in streamlining resource usage, resulting in reduced communication overhead.
\end{abstract}

\end{frontmatter}


\section{Introduction}
Long-term time series forecasting involves predicting future values based on historical data over an extended period. The primary goal is to provide reliable and accurate predictions for multiple timestamps ahead, which can be crucial for planning, decision-making, risk management, and resource allocation in various domains such as finance, energy management, transportation, and environmental monitoring.

Recent advances in deep learning, particularly within Transformer- and LLM-based models~\cite{liu2021pyraformer,zhou2021informer,chen2021autoformer,zhou2022fedformer,nie2022time,Chang2023LLM4TS}, have showcased significant progress in long-term time series forecasting. These models capitalize on the self-attention mechanism and its variants, providing a distinct advantage in capturing and modeling long-range dependencies within sequential data. While LLMs are adept at processing discrete tokens, time series data presents a unique challenge due to its continuous nature. Also, LLMs do not inherently possess the capability to interpret time series patterns, as this knowledge is not typically included in their pre-training. In addition, centralized learning models, including LLM-based methods, often require collecting and storing vast amounts of data in a central server, raising concerns about user privacy and data security. Thus, effectively leveraging LLMs for long-term time series forecasting, while ensuring accuracy and data privacy, highlights the challenges associated with centralized learning models, emphasizing the need for federated learning approaches~\cite{kairouz2021advances}. Federated learning is an emerging machine learning paradigm that enables edge devices to utilize their local datasets to collaboratively train a global model with the help of a central server, while preserving data privacy by keeping the data on the respective devices. At each iteration, the server broadcasts the current global model to the devices for local training, and aggregates the local model updates from the devices to update the global model. Federated learning offers the potential to enhance long-term prediction accuracy~\cite{li2022federated}. However, certain challenges arise from skewed data distributions that may impact the training quality~\cite{li2021survey}. Another challenge in federated learning is systems heterogeneity, which arises from the variation in storage, computational, and communication capabilities influenced by differences in hardware specifications, and power constraints~\cite{kairouz2021advances}. This leads to heightened concerns around stragglers and fault tolerance. Several strategies have been proposed to address this issue involving adaptive optimization techniques that cater to the varying capabilities of devices, optimizing resource usage based on each device's characteristics~\cite{kairouz2021advances}. Additionally, employing techniques like asynchronous updates and gradient compression can reduce communication overhead and accommodate devices with limited connectivity or power~\cite{mcmahan2017communication}. These strategies collectively aim to enhance the efficiency and reliability of federated learning systems in heterogeneous environments. However, these techniques do not significantly enhance model accuracy in scenarios involving optimization and transfer learning, particularly in contexts where data heterogeneity arises from collective settings~\cite{wen2023survey}.

In this paper, we introduce a federated learning framework, dubbed FedTime, for long-term time series forecasting, while preserving data privacy. Local data are used for training individual predictive LLM models on edge devices, allowing for rapid, localized predictions while reducing the load on cloud or central servers. The cloud server plays a dual role in initializing the system and aggregating local LLMs for global updates, a key aspect of federated learning. To optimize the use of limited computational resources on edge devices, we employ parameter-efficient tuning techniques~\cite{hu2021lora,dettmers2023qlora}, which are crucial in minimizing computational and communication overhead, facilitating local training. Our learning framework encompasses training procedures on local edge devices (e.g., EV charging stations) and subsequent aggregation processes on the central server side by leveraging the pre-trained Large Language Model Meta AI (LLaMA-2)~\cite{touvron2023llama}. This LLaMA model is a group of foundational pre-trained LLMs with various sizes (7B, 13B, 33B, and 65B parameters), offering a range of capabilities suited to different needs and computational resources. The key contributions of this paper can be summarized as follows: (i) We introduce a federated large language framework, called FedTime, allowing for collaborative model training across edge devices while preserving data privacy; (ii) we design a model architecture using a two-phase fine-tuning strategy, leveraging quantized and low-ranking adaptation, and aligning the model with time series data via direct preference optimization; and (iii) we demonstrate through extensive experiments the superior performance of FedTime over strong baselines across various forecasting datasets, especially for long-range forecasts. Moreover, we empirically show that FedTime exhibits reduced communication overhead.

\section{Related Work}\label{sec:related_work}
A sizable body of research has been developed to design Transformer-based methods for long-term time series forecasting~\cite{liu2021pyraformer,zhou2021informer,wu2021autoformer,zhou2022fedformer,nie2022time}. Pyraformer~\cite{liu2021pyraformer} presents a pyramidal attention module, leveraging inter-scale and intra-scale connections, which enhances model capability to capture temporal patterns. Informer~\cite{zhou2021informer} design a ProbSparse self-attention mechanism to improve model prediction capacity. Autoformer~\cite{wu2021autoformer} adapts decomposition blocks and auto-correlation mechanisms to help separate the long-term trend information from predicted hidden variables. FEDformer~\cite{zhou2022fedformer} employs fast Fourier transform to decompose sequences, enhancing the extraction of long-term information from the data. PatchTST~\cite{nie2022time} segments time series data into subseries-level patches, which serve as input tokens to a Transformer-based architecture. It also employs channel-independence, where each channel contains a single univariate time series that shares the same embedding and Transformer weights across all series, allowing efficient processing of multiple time series. More recently, various pre-trained LLMs, such as the generative pre-trained Transformer (GPT)~\cite{ouyang2022training} and LLaMA~\cite{touvron2023llama} models rooted in the Transformer architecture, have emerged as powerful methods in generating high-quality outputs in tasks spanning natural language processing~\cite{thoppilan2022lamda,murugesan2023rise,taori2023stanford,zheng2023judging,zhang2023multimodal} and time series forecasting~\cite{Chang2023LLM4TS}, leveraging their ability to discern and comprehend intricate dependencies within data. Chang \textit{et al.} introduce an LLM-based method, called LLM4TS, by adapting the GPT model as a foundational backbone to time series data and then fine-tuning it for time series forecasting tasks. Although our work also involves a two-stage fine-tuning strategy, it differs from LLM4TS in that our FedTime method not only focuses on efficient fine-tuning, but is also tailored for federated learning setups, aiming at long-term time series forecasting. The key strength of FedTime, compared to Transformer- and LLM-based approaches, lies in its ability to allow edge devices to collaborate in model training while keeping data decentralized. This preserves data privacy, a crucial aspect, especially in sensitive domains like healthcare. In addition, FedTime leverages parameter-efficient fine-tuning techniques, enabling it to update only a small number of local model parameters. This efficiency in training and model updates is crucial for real-world applications, especially in resource-constrained edge devices such as EV charging stations.

\section{Method}\label{sec:method}
In this section, we formulate the time series forecasting problem and provide an approach overview. Then, we present the main building blocks of the proposed federated learning framework.

\subsection{Problem Statement and Approach Overview}
\noindent\textbf{Problem Statement.}\quad Time series forecasting refers to the process of predicting future values over a period of time using historical data. Let $\bm{X}_{1:L}=(\bm{x}_{1},\cdots,\bm{x}_{L})^{\T}\in\mathbb{R}^{L\times M}$ be a history sequence of $L$ multivariate time series, where for any time step $t$, each row $\bm{x}_{t}=(x_{t1},\dots,x_{tM})\in\mathbb{R}^{1\times M}$ is a multivariate vector consisting of $M$ variables or channels.

Given a history sequence $\bm{X}_{1:L}$ with look-back window $L$, the goal of multivariate time series forecasting is to predict a sequence $\bm{X}_{L+1:L+T}=(\bm{x}_{L+1},\dots,\bm{x}_{L+T})^{\T}\in\mathbb{R}^{T\times M}$ for the future $T$ timesteps. To this end, we introduce a federated pre-trained large language model by leveraging channel independence and patching~\cite{nie2022time}, as well as supervised and downstream fine-tuning strategies~\cite{Chang2023LLM4TS}.

\medskip\noindent\textbf{Approach Overview.}\quad  The proposed federated learning framework comprises two main components: local training operations conducted on the client side and aggregation operations carried out on the server side. These components collaborate to facilitate efficient training. Prior to the learning process, we apply K-means clustering as a pre-processing step to group similar edge devices or clients (i.e., based on cluster size and performance). Clustering not only helps reduce the impact of biased prediction~\cite{Saputra2019FEDL}, but also ensures that sensitive data remains localized within clusters, reducing the risk of data exposure during model training. Then, we employ a pre-trained LLaMA model~\cite{touvron2023llama} as the foundational backbone to leverage the inherent capabilities of LLMs in capturing data patterns, thereby facilitating its application to time series forecasting within our federated learning framework. To improve the global model convergence rate, we use federated averaging with adaptive learning rates using Adam optimizer~\cite{mills2019communication}, which dynamically adjusts the learning rates for each local device based on the model updates. For each cluster, a global model is produced by aggregating the model updates from the cluster members. We also employ parameter-efficient fine-tuning techniques~\cite{dettmers2023qlora}, which offer an efficient way to adapt pre-trained LLMs to diverse downstream tasks by focusing on updating only a small number of model parameters, and hence transmit updates faster, leading to quicker convergence and reduced communication overhead. While our proposed approach builds upon these existing techniques, its technical novelties lie in how these techniques are integrated and adapted to address specific challenges in long-term time series forecasting. In addition to proposing a novel two-phase fine-tuning strategy that leverages quantized and low-ranking adaptation techniques, along with direct preference optimization, to align the model with time series data effectively, our federated learning framework streamlines resource usage by minimizing computational and communication overhead, thereby improving the scalability and efficiency of the training process.

\subsection{Proposed Learning Framework}
\begin{figure}[!htb]
\centering
\includegraphics[scale=0.46]{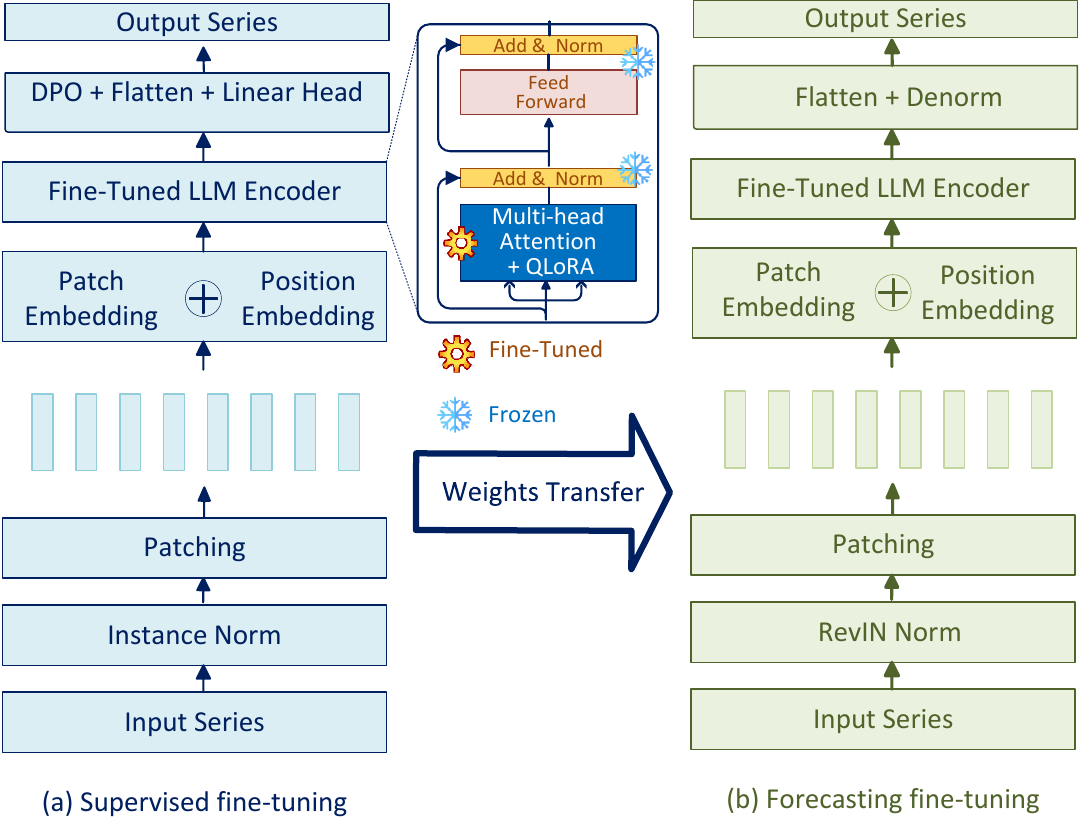}
\caption{\textbf{FedTime Architecture}. (a) Supervised fine-tuning with direct preference optimization (DPO) integration for backbone model alignment. (b) Forecasting fine-tuning using reversible instance normalization (RevIN).}
\label{Fig:FedTime_arch}
\end{figure}

We present federated learning framework, designed to overcome the classical challenges within a centralized training environment, such as network congestion, low bandwidth, latency, and privacy. The proposed federated architecture is built on the LLaMA structure, enabling parallel feature learning across various time series forecasting benchmarks. Figure~\ref{Fig:FedTime_arch} depicts the main building blocks of our learning framework, including time series data normalization, patching and positional encodings, LLM encoder and robust fine-tuning strategies. Our two-phase approach consists of supervised fine-tuning, followed by forecasting fine-tuning.

\medskip\noindent\textbf{Channel Independence.}\quad Channel independence refers to a strategy where the features or variables of a multivariate time series are treated separately~\cite{nie2022time}, emphasizing independence rather than mixing or amalgamating information from different channels or variables. This approach aims to retain the distinct characteristics of each variable without merging them together. Specifically, the input time series $\bm{X}_{1:L}=(\bm{x}_{1},\dots,\bm{x}_{L})^{\T}$ is split into $M$ univariate series $\bm{x}^{(i)}=(x_{1}^{(i)},\dots,x_{L}^{(i)})^{\T}\in\mathbb{R}^{L}$, where $\bm{x}^{(i)}$ is the $i$th column of $\bm{X}_{1:L}$. Each of these univariate series is fed into the model backbone. Our proposed federated large language model takes $\bm{x}^{(i)}$ as input and returns a $T$-dimensional vector of predictions $\hat{\bm{x}}^{(i)}=(\hat{x}_{L+1}^{(i)},\dots,\hat{x}_{L+T}^{(i)})^{\T}$.

\medskip\noindent\textbf{Normalization.}\quad In the supervised fine-tuning phase, we employ instance normalization before patching to normalize each data instance $\bm{x}^{(i)}$ with zero mean and unit standard deviation, which are then added back to the output prediction. In the subsequent forecasting fine-tuning phase, each input series is normalized using the reversible instance normalization (RevIN) technique~\cite{Kim2022RevIN}, which addresses challenges related to shifts in data distributions over time. RevIN consists of two main steps: normalization and denormalization. In the first step, the input undergoes normalization to standardize its distribution in terms of mean and variance. After the model generates output sequences, RevIN reverses the normalization process by denormalizing these outputs.

\medskip\noindent\textbf{Patching.}\quad We adopt the patching strategy from~\cite{nie2022time}, which involves dividing the sequential data into smaller patches. Specifically, each input series $\bm{x}^{(i)}\in\mathbb{R}^{L}$ is partitioned into patches, resulting in a sequence of patches $\bm{X}^{(i)}_{p}\in\mathbb{R}^{N\times P}$, where $P$ is the patch length and $N$ is the resulting number of patches, which also serves as the input sequence length for the Transformer encoder. The benefits of patching include better preservation of local semantic information in patches, computational and memory efficiency, and an expanded historical view.

\medskip\noindent\textbf{Patch and Position Embeddings.}\quad Patch embeddings are obtained by mapping each patch in the sequence of patches to $D$ dimensions with a trainable linear projection $\bm{W}_{p}\in\mathbb{R}^{P\times D}$. Position embeddings are added to patch embeddings via a learnable position encoding matrix $\bm{W}_{\text{pos}}\in\mathbb{R}^{N\times D}$ to retain positional information of the patches. Summing patch and position embeddings yields an embedding $\bm{X}^{(i)}_{d}\in\mathbb{R}^{N\times D}$ given by
\begin{equation}
\bm{X}^{(i)}_{d} = \bm{X}^{(i)}_{p}\bm{W}_{p} + \bm{W}_{\text{pos}},
\end{equation}
which serves as input to the multi-head self-attention mechanism of the Transformer encoder.

\medskip\noindent\textbf{LLM Encoder.}\quad We employ the lightweight LLaMA2-7B model as our LLM encoder. This LLaMA model is fundamentally built on the Transformer architecture, distinguished by its 7 billion parameters, and benefits from a comprehensive training on an expansive corpus containing 2 trillion tokens. At their core, Transformers rely on self-attention, which operates on the matrix $\bm{X}^{(i)}_{d}$ of tokens (i.e., patch embeddings with added positional encodings) by first linearly projecting it into a query matrix $\bm{Q}^{(i)}$, a key matrix $\bm{K}^{(i)}$ and a value matrix $\bm{V}^{(i)}$ as follows:
\begin{equation}
\bm{Q}^{(i)}=\bm{X}^{(i)}_{d}\bm{W}_{q},\,\, \bm{K}^{(i)}=\bm{X}^{(i)}_{d}\bm{W}_{k},\,\, \bm{V}^{(i)}=\bm{X}^{(i)}_{d}\bm{W}_{v},
\end{equation}
where $\bm{W}_{q}\in\mathbb{R}^{D\times d_{k}}$, $\bm{W}_{k}\in\mathbb{R}^{D\times d_{k}}$ and $\bm{W}_{v}\in\mathbb{R}^{D\times d_{k}}$ are learnable weight matrices, and $d_k$ is the projection dimension. Then, the self-attention (SA) output is a $N\times d_{k}$ matrix defined as the weighted sum of the value vectors for each token
\begin{equation}
\mathsf{SA}(\bm{X}^{(i)}_{d})=\mathsf{softmax}\left(\frac{\bm{Q}^{(i)}\bm{K}^{{(i)}^{\T}}}{\sqrt{d_k}}\right)
\bm{V}^{(i)},
\end{equation}
where the weights are attention scores computed as the dot product between the query and key vectors, scaled by the square root of the projection dimension, and followed by softmax applied row-wise.

In the multi-head attention mechanism (MSA), multiple self-attention operations are performed in parallel, each with its own set of learnable weight matrices. For simplicity, we assume $d_k=D/h$, where $h$ is the number of attention heads. The outputs of $h$ heads are concatenated and linearly transformed using a learnable weight matrix $\bm{W}_{o}\in\mathbb{R}^{D\times D}$ to obtain the MSA output:
\begin{equation}
\mathsf{MSA}(\bm{X}^{(i)}_{d})=\mathsf{Concat}(\bm{Y}^{(i)}_1,\dots,\bm{Y}^{(i)}_h)\bm{W}_{o}\in\mathbb{R}^{N\times D},
\end{equation}
where $\bm{Y}^{(i)}_{j}=\mathsf{SA}_{j}(\bm{X}^{(i)}_{d})\in\mathbb{R}^{N\times \frac{D}{h}}$ is the output of the $j$-th attention head.

As shown in Figure~\ref{Fig:FedTime_arch}, the Transformer encoder architecture is comprised of a series of alternating layers, each containing an MSA mechanism and a feedforward neural network (FFN) block with SwiGLU activation function~\cite{shazeer2020glu}. Before each block, layer normalization is applied, and after each block, residual connections are employed. Hence, the resulting output of the Transformer encoder is $\bm{Z}^{(i)}_{d}\in\mathbb{R}^{N\times D}$, which subsequently undergoes a flatten layer with linear head to produce the model predictions $\hat{\bm{x}}^{(i)}=(\hat{x}_{L+1}^{(i)},\cdots,\hat{x}_{L+T}^{(i)})^{\T}$.

\medskip\noindent\textbf{Model Fine-Tuning.}\quad Parameter efficient fine-tuning (PEFT) methods offer an efficient way to adapt pre-trained LLMs to diverse downstream tasks without the need to fine-tune the entire set of model parameters. To fine-tune our pre-trained LLM, we employ quantized and low-rank adaptation (QLoRA)~\cite{dettmers2023qlora}, a PEFT technique that aims to minimize the number of parameters updated during fine-tuning while preserving or improving performance on downstream tasks. QLoRA builds upon LoRA~\cite{hu2021lora}, which freezes the weight matrices of the linear projection layers within the self-attention mechanism of the Transformer encoder and represents each of their update matrices as a product of two trainable low-rank matrices. This low-rank matrix decomposition significantly reduces the number of trainable parameters, making the model more efficient for downstream tasks. In addition to low-rank decomposition, QLoRA also incorporates quantization techniques that reduce the precision of the weight values, thereby further decreasing the model's memory footprint and computational requirements without substantially sacrificing performance. With QLoRA, only 1.2\% of the model's parameters are considered trainable, whereas using LoRA increases this percentage to 1.5\% \cite{Chang2023LLM4TS}.

\medskip\noindent\textbf{Model Alignment.}\quad Supervised fine-tuning is used as a fundamental precursor for subsequent forecasting fine-tuning of the LLaMA model. To enhance the alignment of our model with time series data after the supervised fine-tuning phase, we integrate direct preference optimization (DPO)~\cite{rafailov2023direct} into our fine-tuning process. DPO, which involves the analysis of 10K comparison data pairs from the UltraFeedback dataset~\cite{cui2023ultrafeedback}, is applied post-supervised fine-tuning to enhance model performance for specific data points. It optimizes its objective function, producing the optimal policy to an implicit reward function that is the best fit to the preference data. We use DPO to capture any change of variables, ensuring a more effective adaptation of the LLaMA model to the intricacies of time series forecasting. Finally, we transfer the updated weights of the backbone model to the forecasting fine-tuning phase.

\medskip\noindent\textbf{Loss Function.}\quad The parameters of our FedTime model are learned by minimizing the training objective, which is the mean squared error (MSE) defined as
\begin{equation}
\mathcal{L}=\frac{1}{MT}\sum_{i=1}^{M}\sum_{\tau=L+1}^{L+T}\big\Vert \bm{x}_{\tau}^{(i)} - \hat{\bm{x}}_{\tau}^{(i)}\big\Vert^2,
\end{equation}
where $\bm{x}_{\tau}^{(i)}$ and $\hat{\bm{x}}_{\tau}^{(i)}$ are the ground-truth and prediction, respectively, $\tau\in\{L+1,\dots,L+T\}$, $L$ is the look-back window, $T$ is the prediction length/horizon (i.e., future timesteps), and $M$ is the number of time series variables (i.e., number of features or channels).

\medskip\noindent\textbf{Algorithm.}\quad The main algorithmic steps of our federated learning framework are summarized in Algorithm~\ref{Algo:fed-llm}. Prior to the learning process, K-means clustering is used as a pre-processing step to group edge devices into clusters. In practice, K-means clustering may occur in scenarios where a large number of edge devices are deployed in geographically dispersed locations, such as smart city infrastructures, industrial IoT deployments, or distributed sensor networks.

The FedTime algorithm initializes a global model, distributes it to selected edge devices for local training, aggregates the updated models from the participating devices at the central server, and iterates this process for multiple rounds, allowing the global model to gradually improve by aggregating insights from diverse edge devices while preserving data privacy. By training cluster-specific models, FedTime can capture patterns specific to different subsets of edge devices. Moreover, instead of transmitting updates from each individual edge device to the central server, FedTime aggregates updates from devices within the same cluster. This helps reduce the overall communication overhead by consolidating updates from devices with similar data distributions.

\begin{algorithm}[!htb]
\caption{FedTime: Federated LLM Algorithm}\label{Algo:fed-llm}
\label{alg:federated_clustering}
\begin{algorithmic}[1]
\STATE \textbf{Input:} Global model parameters $\bg{\theta}$; number of edge devices $S$; number of clusters $K$; local datasets $\mathcal{D}_1, \dots, \mathcal{D}_S$
\STATE \textbf{Output:} Vector of aggregated global cluster-specific models $\bg{\theta}_{K}$
\STATE Apply K-means algorithm to split devices into $K$ clusters $\mathcal{C}=\{c_{1},c_{2},\dots,c_{K}\}$ based on similar attributes.
\STATE Initialize global cluster model parameters $\bg{\theta}_K$
\FOR{each cluster $c$ in $\mathcal{C}$}
    \FOR{each round  $r=1,2,\dots$}
    \WHILE{model $\bg{\theta}_K$ has not converged yet}
    \FOR{each edge device $s\in\{1,\dots,S\}$ \textbf{in parallel}}
        \STATE $\bg{\theta}_{K} \leftarrow$ \textbf{UpdateDevice}($\mathcal{D}_s, \bg{\theta}_c$)
    \ENDFOR
    \ENDWHILE
    \STATE Server aggregates final global models from cluster-specific models using weighted averaging
    $$\bg{\theta}_c = \frac{\sum_{s} w_{s,c} \bg{\theta}_K}{\sum_{s} w_{s,c}}$$
    where $w_{s,c}$ is the weight of device $s$ in cluster $c$.
    \ENDFOR
    \STATE Send global cluster-specific model $\bg{\theta}_c$ to local device $s$
\ENDFOR

\STATE \textbf{UpdateDevice($\mathcal{D}_s, \bg{\theta}$)}
\FOR{each cluster $c$ in $\mathcal{C}$}
    \FOR{each local device $s$ in cluster $c$}
        \STATE Receive global model parameters $\bg{\theta}$ from the server
        \STATE Train local model using local data $\mathcal{D}_s$
        \STATE Update local model parameters
        $$\bg{\theta}\leftarrow \bg{\theta} - \eta\nabla\mathcal{L}(\bg{\theta}; \mathcal{D}_s)$$
        where $\eta$ is the learning rate and $\mathcal{L}(\bg{\theta}; \mathcal{D}_s)$ is the MSE local loss.
        \STATE Send local model parameters $\bg{\theta}$ to the server
    \ENDFOR
\ENDFOR
\STATE \textbf{Return:} $\bg{\theta}_K$
\end{algorithmic}
\end{algorithm}

\section{Experiments}\label{sec:experiments}
In this section, we present experimental evaluation of the proposed federated learning approach by comparing it with recent state-of-the-art methods.

\subsection{Experimental Setup}
\noindent{\textbf{Datasets.}}\quad We conduct a comprehensive evaluation of our FedTime model on various benchmark datasets: Weather\footnote{https://www.bgc-jena.mpg.de/wetter}, Traffic\footnote{http://pems.dot.ca.gov}, Electricity\footnote{https://archive.ics.uci.edu/ml/datasets/ElectricityLoadDiagrams20112014}, and four ETT datasets (ETTh1, ETTh2, ETTm1, ETTm2)~\cite{wu2021autoformer}. Dataset statistics are summarized in Table~\ref{Tab:datasets}: (1) \textbf{Weather} is recorded at 10-minute intervals for the entire year of 2020, and contains 21 meteorological indicators, such as air temperature and humidity. (2) \textbf{Traffic} is a collection of hourly data from California Department of Transportation, and describes the road occupancy rates measured by different sensors on San Francisco Bay area freeways.  (3) \textbf{Electricity} contains the hourly electricity consumption of 321 customers from 2012 to 2014. (4) \textbf{ETT} contains data collected from electricity transformers, including load and oil temperature that are recorded every 1 hour for ETTh datasets and every 15 minutes for ETTm datasets between July 2016 and July 2018.

\begin{table}[!htb]
\caption{Summary statistics of benchmark datasets.}
\bigskip
\centering
\begin{tabular}{lrrr}
\toprule
Dataset & Features & Timesteps & Granularity \\
\midrule
Weather & 21 & 52,696 & 10 min \\
Traffic & 862 & 17,544 & 1 hour \\
Electricity & 321 & 26,304 & 1 hour \\
ETTh1 & 7 & 17,420 & 1 hour \\
ETTh2 & 7 & 17,420 & 1 hour \\
ETTm1 & 7 & 69,680 & 15 min \\
ETTm2 & 7 & 69,680 & 15 min \\
\bottomrule
\end{tabular}
\label{Tab:datasets}
\end{table}

\medskip\noindent{\textbf{Baseline Methods.}} We evaluate the performance of our FedTime model against recent state-of-the-art methods, including Pyraformer~\cite{liu2021pyraformer}, Informer~\cite{zhou2021informer}, Autoformer ~\cite{chen2021autoformer}, FEDformer~\cite{zhou2022fedformer}, DLinear~\cite{zeng2023transformers}, N-HiTS~\cite{challu2023nhits}, PatchTST~\cite{nie2022time}, TiDE~\cite{das2023long}, and LLM4TS~\cite{Chang2023LLM4TS}.

\medskip\noindent\textbf{Evaluation Metrics.}\quad We use the mean squared error (MSE) and mean absolute error (MAE) as metrics for evaluating both long-term time series forecasting. MSE measures the average squared difference between the predicted and actual values. A lower MSE indicates that the predictions are closer to the actual values on average. On the other hand, MAE measures the average absolute difference between the predicted and actual values. Similar to MSE, a lower MAE indicates better predictive accuracy, with predictions closer to the actual values. We report the performance at various prediction horizons and look-back windows.

\medskip\noindent{\textbf{Implementation Details.}}\quad In our federated learning setup, we assume the presence of 555 eligible edge devices for training. All experiments are carried out on a GPU server with an AMD EPYC 7502 processor, an NVIDIA RTX A6000 48GB GDDR6 and 512GB memory. The algorithms are implemented in PyTorch and PySyft to simulate a federated learning environment. For hyper-parameters optimization, we apply a grid search on a batch size of range \{128, 256, 512, 1024\}, a learning rate of range \{0.01, 0.001, 0.0001\}, and a local training round of range \{40, 80, 200\}. Using hyper-parameter tuning, the batch size and learning rate are set to 512 and 0.001, respectively. In all experiments, we split the datasets into 80\% training and 20\% testing.  To update QLoRA parameters, we employ FedAdam~\cite{reddi2021fedAdam}, an adaptive optimization method that helps improve the convergence of federated learning.

\subsection{Results and Analysis}
\noindent\textbf{Long-Term Time Series Forecasting.}\quad We evaluate the forecasting performance of our FedTime model against strong baseline methods on seven forecasting benchmark datasets, and the results are reported in Table~\ref{tab:ComparativeResults}. For Transformer-based baselines, the values of the MSE and MAE error metrics are taken from~\cite{nie2022time}. The best results are shown in bold, and the second best results are underlined. As can be seen, FedTime consistently outperforms the baselines on most datasets, particularly for long-term prediction horizons. In terms of MSE, Table~\ref{tab:ComparativeResults} reveals that FedTime performs better than LLM4TS in 24 out of 28 evaluations spanning seven datasets and four forecast horizons, improving upon this best performing baseline by relative error reductions of 15.56\% and 20\% on the Traffic and Electricity datasets, respectively, for $T=720$, while maintaining a smaller number of learnable parameters compared to LLM4TS. Similarly, our method outperforms LLM4TS in 25 out of 28 MAE evaluations, yielding relative error reductions of 10.98\% and 7.36\% on the ETTm1 and ETTm2 datasets, respectively, for $T=720$. Moreover, FedTime performs better or on par with the best performing LLM4TS baseline on the Weather dataset across all the prediction horizons. It is worth pointing out that MAE is more interpretable than MSE since it directly represents the average magnitude of prediction errors. A low MAE implies that FedTime provides accurate estimates of forecasting time series, thus demonstrating the effectiveness of the our framework.

\begin{table*}[!htb]
\caption{Long-term time series forecasting results for various prediction lengths $T\in\{96, 192, 336, 720\}$. The best results are in \textbf{bold}, while the second-best are \underline{underlined}.}
\bigskip
\small 
\setlength\tabcolsep{3.7pt} 
\centering
\begin{tabular}{l*{16}{c}}
\toprule
Method & \multicolumn{2}{c}{Autoformer} & \multicolumn{2}{c}{FEDformer} & \multicolumn{2}{c}{DLinear} & \multicolumn{2}{c}{N-HiTS} & \multicolumn{2}{c}{PatchTST/64} & \multicolumn{2}{c}{TiDE} &\multicolumn{2}{c}{LLM4TS} &\multicolumn{2}{c}{\textbf{FedTime}}\\
\cmidrule(lr){2-3} \cmidrule(lr){4-5} \cmidrule(lr){6-7} \cmidrule(lr){8-9} \cmidrule(lr){10-11} \cmidrule(lr){12-13} \cmidrule(lr){14-15} \cmidrule(lr){16-17}
Metric & MSE & MAE & MSE & MAE & MSE & MAE & MSE & MAE & MSE & MAE & MSE & MAE & MSE & MAE & MSE & MAE \\
\midrule
\multicolumn{1}{r|}{\parbox[t]{2mm}{\multirow{4}{*}{\rotatebox[origin=c]{90}{Weather}}}\quad 96} &  0.249 & 0.329 & 0.238 & 0.314 & 0.176 & 0.237 & 0.158 & 0.195 & \underline{0.149} & 0.198 & 0.166 & 0.222 & \textbf{0.147} & \textbf{0.196} &0.150 & \underline{0.194}\\
\multicolumn{1}{r|}{\quad 192} &  0.325 & 0.370 & 0.275 & 0.329 & 0.220 & 0.282 & 0.211 & 0.247 & 0.194 & 0.241 & 0.209 & 0.263 & \underline{0.191} & \underline{0.238} & \textbf{0.180} & \textbf{0.217} \\
\multicolumn{1}{r|}{\quad 336} & 0.351 & 0.391 & 0.339 & 0.377 & 0.265 & 0.319 & 0.274 & 0.300 & 0.245 & 0.282 & 0.254 & 0.301 & \underline{0.241} & \underline{0.227} &\textbf{0.209} & \textbf{0.227} \\
\multicolumn{1}{r|}{\quad 720} & 0.415 & 0.426 & 0.389 & 0.409 & 0.323 & 0.362 & 0.401 & 0.413 & 0.314 & 0.334 & 0.313 & 0.340 &  \textbf{0.313}& \textbf{0.329} & \underline{0.318} & \underline{0.335} \\
\midrule
\multicolumn{1}{r|}{\parbox[t]{2mm}{\multirow{4}{*}{\rotatebox[origin=c]{90}{Traffic}}}\quad 96} &  0.597 & 0.371 & 0.576 & 0.359 & 0.410 & 0.282 & 0.402 & 0.282 & 0.360 & \underline{0.249} & \underline{0.336} & 0.253 & 0.372 & 0.259 & \textbf{0.328} & \textbf{0.215} \\
\multicolumn{1}{r|}{\quad 192} &  0.607 & 0.382 & 0.610 & 0.380 & 0.423 & 0.287 & 0.420 & 0.297 & 0.379 & \underline{0.256} & \underline{0.346} & 0.257  & 0.391 & 0.265 & \textbf{0.339} & \textbf{0.233} \\
\multicolumn{1}{r|}{\quad 336} &  0.623 & 0.387 & 0.608 & 0.375 & 0.436 & 0.296 & 0.448 & 0.313 & 0.392 & 0.264 & \underline{0.355} & \textbf{0.260} & 0.405 & \underline{0.275} & \textbf{0.347} & 0.288 \\
\multicolumn{1}{r|}{\quad 720} & 0.639 & 0.395 & 0.621 & 0.375 & 0.466 & 0.315 & 0.539 & 0.353 & 0.432 & 0.286 & \underline{0.386} & \underline{0.273} & 0.437 & 0.292 & \textbf{0.369} & \textbf{0.239} \\
\midrule
\multicolumn{1}{r|}{\parbox[t]{2mm}{\multirow{4}{*}{\rotatebox[origin=c]{90}{Electricity}}}\quad 96} &  0.196 & 0.304 & 0.186 & 0.311 & 0.140 & 0.237 & 0.147 & 0.249 & 0.129 & 0.222 & 0.132 & 0.229 & \underline{0.128} & \underline{0.223} & \textbf{0.118} & \textbf{0.158} \\
\multicolumn{1}{r|}{\quad 192} &  0.211 & 0.324 & 0.197 & 0.311 & 0.153 & 0.249 & 0.167 & 0.269 & 0.147 & \underline{0.240} & \underline{0.147} & 0.243 & \underline{0.146} & 0.242 & \textbf{0.133} & \textbf{0.315} \\
\multicolumn{1}{r|}{\quad 336} & 0.214 & 0.327 & 0.213 & 0.314 & 0.169 & 0.267 & 0.186 & 0.290 & 0.163 & 0.259 & \underline{0.161} & 0.261 & 0.163 & \underline{0.258} & \textbf{0.148} & \textbf{0.255} \\
\multicolumn{1}{r|}{\quad 720} &  0.236 & 0.342 & 0.233 & 0.344 & 0.203 & 0.301 & 0.243 & 0.340 & 0.197 & 0.290 & \underline{0.196} & 0.294 & 0.220 & \underline{0.292} & \textbf{0.176} & \textbf{0.288}  \\
\midrule
\multicolumn{1}{r|}{\parbox[t]{2mm}{\multirow{4}{*}{\rotatebox[origin=c]{90}{ETTh1}}}\quad 96} &  0.435 & 0.446 & 0.376 & 0.415 & 0.375 & 0.399 & 0.378 & 0.436 & 0.379 & 0.401 & \underline{0.375} & \underline{0.398} & \textbf{0.371} & \textbf{0.394} & 0.392 & 0.402 \\
\multicolumn{1}{r|}{\quad 192} & 0.456 & 0.457 & 0.423 & 0.446 & 0.405 & 0.420 & 0.427 & 0.436 & 0.413 & 0.429 & 0.412 & 0.422 & \textbf{0.403} & \underline{0.412} & \underline{0.404} & \textbf{0.360} \\
\multicolumn{1}{r|}{\quad 336} & 0.486 & 0.487 & 0.444 & 0.462 & 0.439 & 0.443 & 0.458 & 0.484 & 0.435 & 0.436 & 0.435 & 0.433 & \underline{0.428} & \underline{0.422} &\textbf{0.425} & \textbf{0.421} \\
\multicolumn{1}{r|}{\quad 720} & 0.515 & 0.517 & 0.469 & 0.492 & 0.472 & 0.490 & 0.472 & 0.551 & \underline{0.446} & 0.464 & 0.454 & 0.465 & \underline{0.422} & \underline{0.444} & \textbf{0.414} & \textbf{0.436} \\
\midrule
\multicolumn{1}{r|}{\parbox[t]{2mm}{\multirow{4}{*}{\rotatebox[origin=c]{90}{ETTh2}}}\quad 96} & 0.332 & 0.368 & 0.332 & 0.374 & 0.289 & 0.353 & 0.274 & 0.345 & 0.274 & 0.337 & 0.270 & 0.336 & \underline{0.269} & \underline{0.332} & \textbf{0.249} & \textbf{0.303} \\
\multicolumn{1}{r|}{\quad 192} & 0.426 & 0.434 & 0.407 & 0.446 & 0.383 & 0.418 & 0.353 & 0.401 & 0.338 & \underline{0.376} & 0.332 & 0.380 & \underline{0.328} & 0.377 & \textbf{0.318} & \textbf{0.343} \\
\multicolumn{1}{r|}{\quad 336} & 0.477 & 0.479 & 0.400 & 0.447 & 0.448 & 0.465 & 0.382 & 0.425 & 0.363 & 0.397 & 0.360 & 0.407 & \underline{0.353} & \underline{0.396} & \textbf{0.344} & \textbf{0.368} \\
\multicolumn{1}{r|}{\quad 720} & 0.453 & 0.490 & 0.412 & 0.469 & 0.605 & 0.551 & 0.625 & 0.557 & 0.393 & 0.430 & 0.419 & 0.451 & \underline{0.383} &  \underline{0.425}& \textbf{0.376} & \textbf{0.428} \\
\midrule
\multicolumn{1}{r|}{\parbox[t]{2mm}{\multirow{4}{*}{\rotatebox[origin=c]{90}{ETTm1}}}\quad 96} &  0.510 & 0.492 & 0.326 & 0.390 & 0.299 & 0.343 & 0.302 & 0.350 & 0.293 & 0.346 & 0.306 & 0.349 & \underline{0.285} & \underline{0.343} &\textbf{0.283} & \textbf{0.339} \\
\multicolumn{1}{r|}{\quad 192} &  0.514 & 0.495 & 0.365 & 0.415 & 0.335 & 0.365 & 0.347 & 0.383 & 0.333 & 0.370 & 0.335 & \underline{0.366} & \underline{0.324} & \underline{0.366} & \textbf{0.301} & \textbf{0.338}\\
\multicolumn{1}{r|}{\quad 336} & 0.510 & 0.492 & 0.392 & 0.425 & 0.369 & 0.386 & 0.369 & 0.402 & 0.369 & 0.392 & 0.364 & \underline{0.384} & \underline{0.353} & 0.385 &\textbf{0.319} & \textbf{0.365} \\
\multicolumn{1}{r|}{\quad 720} &  0.527 & 0.493 & 0.446 & 0.458 & 0.425 & 0.421 & 0.431 & 0.441 & 0.416 & 0.420 & 0.413 & \underline{0.413} & \underline{0.408} & 0.419 & \textbf{0.328} & \textbf{0.373} \\
\midrule
\multicolumn{1}{r|}{\parbox[t]{2mm}{\multirow{4}{*}{\rotatebox[origin=c]{90}{ETTm2}}}\quad 96} & 0.205 & 0.293 & 0.180 & 0.271 & 0.167 & 0.260 & 0.176 & 0.255 & 0.166 & 0.256 & \underline{0.161} & \underline{0.251} & 0.165 & 0.254 & \textbf{0.148} & \textbf{0.227}\\
\multicolumn{1}{r|}{\quad 192} & 0.278 & 0.336 & 0.252 & 0.318 & 0.224 & 0.303 & 0.245 & 0.305 & 0.223 & 0.296 & \underline{0.215} & \underline{0.289} & 0.225 & 0.292 & \textbf{0.197} & \textbf{0.261}  \\
\multicolumn{1}{r|}{\quad 336} & 0.343 & 0.379 & 0.324 & 0.364 & 0.281 & 0.342 & 0.295 & 0.346 & 0.274 & 0.329 & \underline{0.267} & \underline{0.326} & 0.268 & 0.326 & \textbf{0.219} & \textbf{0.279} \\
\multicolumn{1}{r|}{\quad 720} &  0.414 & 0.419 & 0.410 & 0.420 & 0.397 & 0.421 & 0.401 & 0.413 & 0.362 & 0.385 & \underline{0.352} & 0.383 & 0.356& \underline{0.380} & \textbf{0.305} & \textbf{0.352}\\
\bottomrule
\end{tabular}
\label{tab:ComparativeResults}
\end{table*}

\begin{figure*}[!htb]
\centering
\includegraphics[scale=0.33]{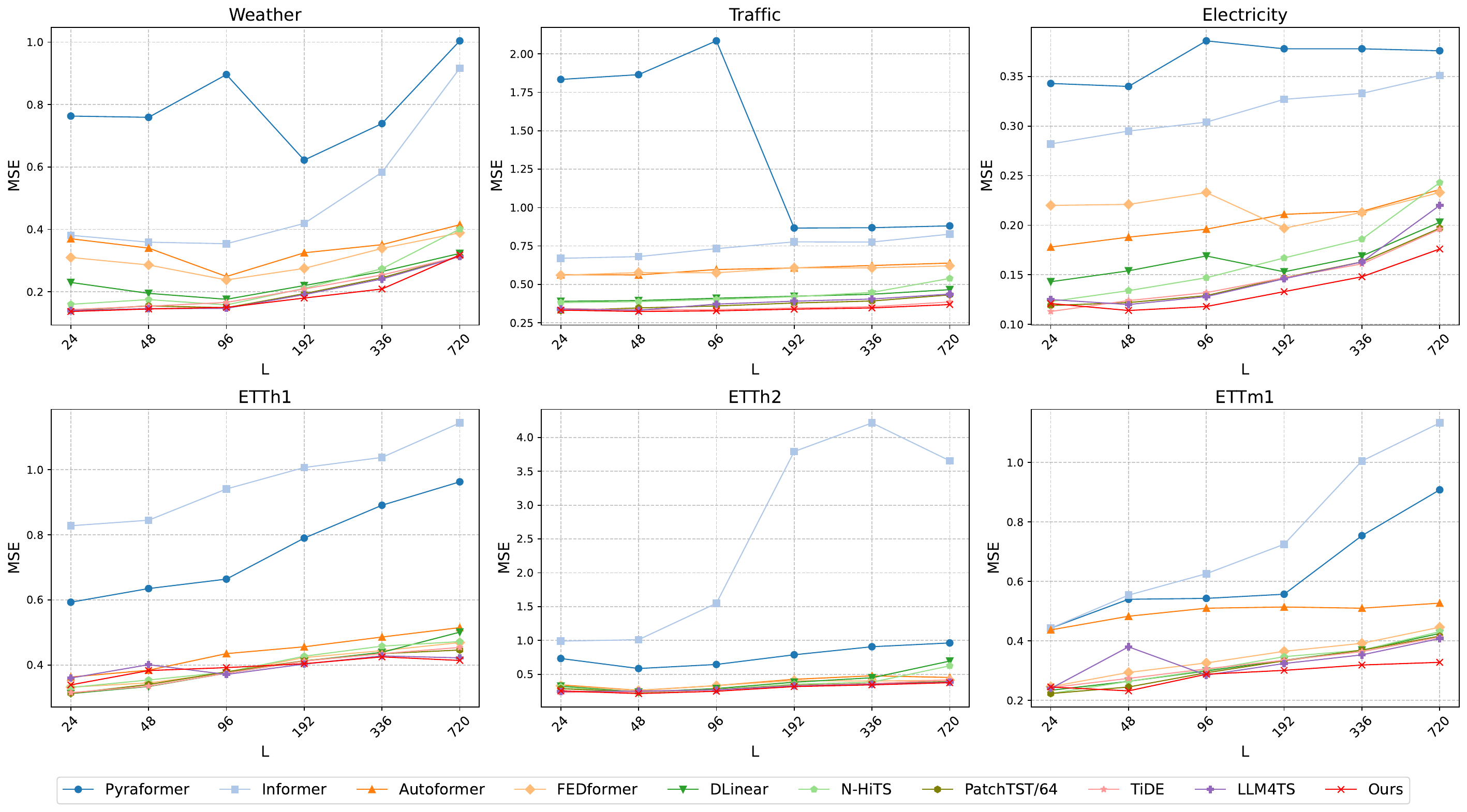}
\caption{Long-term forecasting performance with varying look-back window lengths $L\in\{24, 48, 96, 192, 336, 720\}$. We set the prediction horizon to $T=720$.}
\label{Fig:Multiple_MSE_Curve}
\end{figure*}

In Figure~\ref{Fig:Multiple_MSE_Curve}, we report the visual results in terms of MSE on six forecasting datasets for various values of the look-back window length $L$ ranging from 24 to 720. As can be seen, FedTime exhibits consistent and superior performance, especially for long look-back windows. On the Traffic dataset, it is evident that FedTime holds a competitive edge, consistently outperforming Transformer-based models such as Informer, Autoformer and FEDformer. On the Weather dataset, FedTime stands out, especially between the look-back window lengths of 192 and 336. On the ETTh1 and ETTh2 satasets, FedTime's performance peaks within the intermediate look-back window span of 48 to 192. Overall, FedTime yields superior performance over the baselines across the various forecasting datasets, especially for long-range look-back scenarios.

\begin{figure}[!htb]
\setlength\tabcolsep{1.5pt}
\centering
\begin{tabular}{cc}
\includegraphics[scale=0.21]{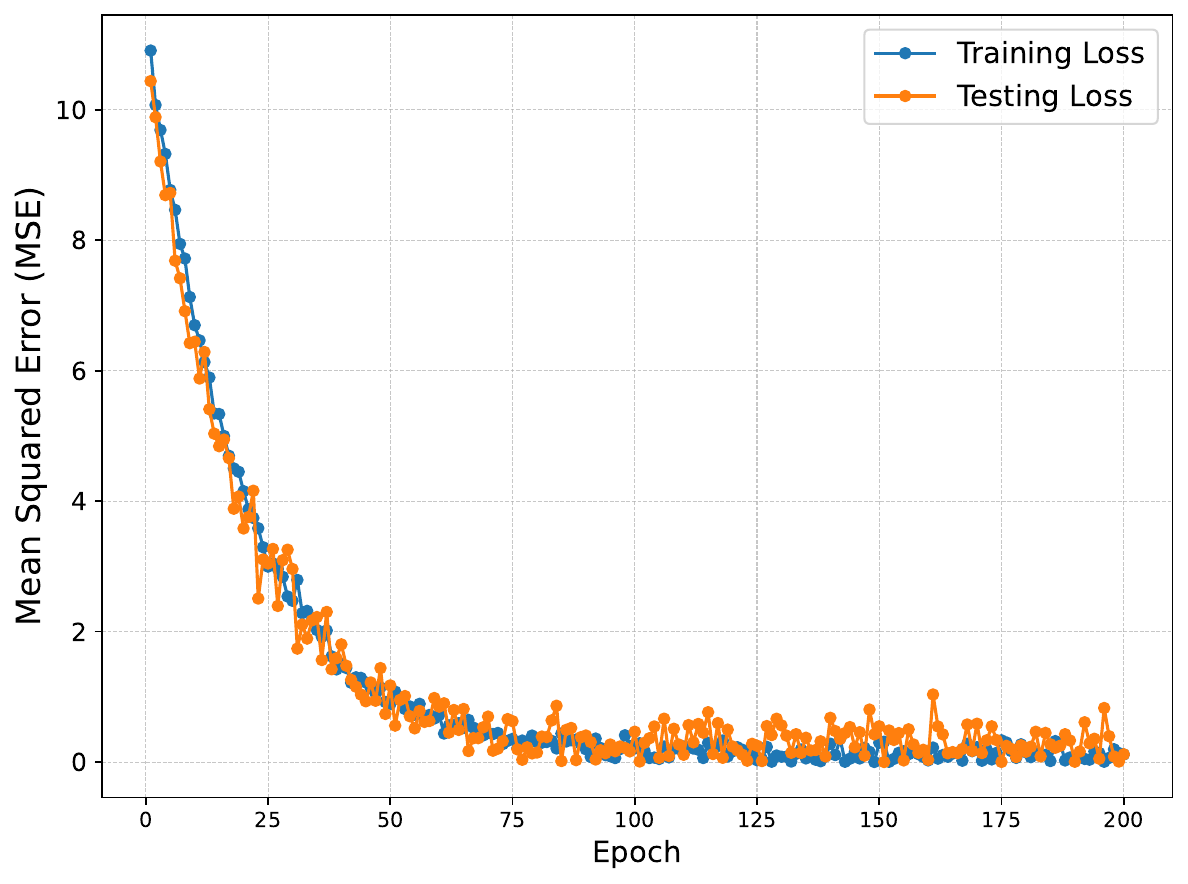} & \includegraphics[scale=0.21]{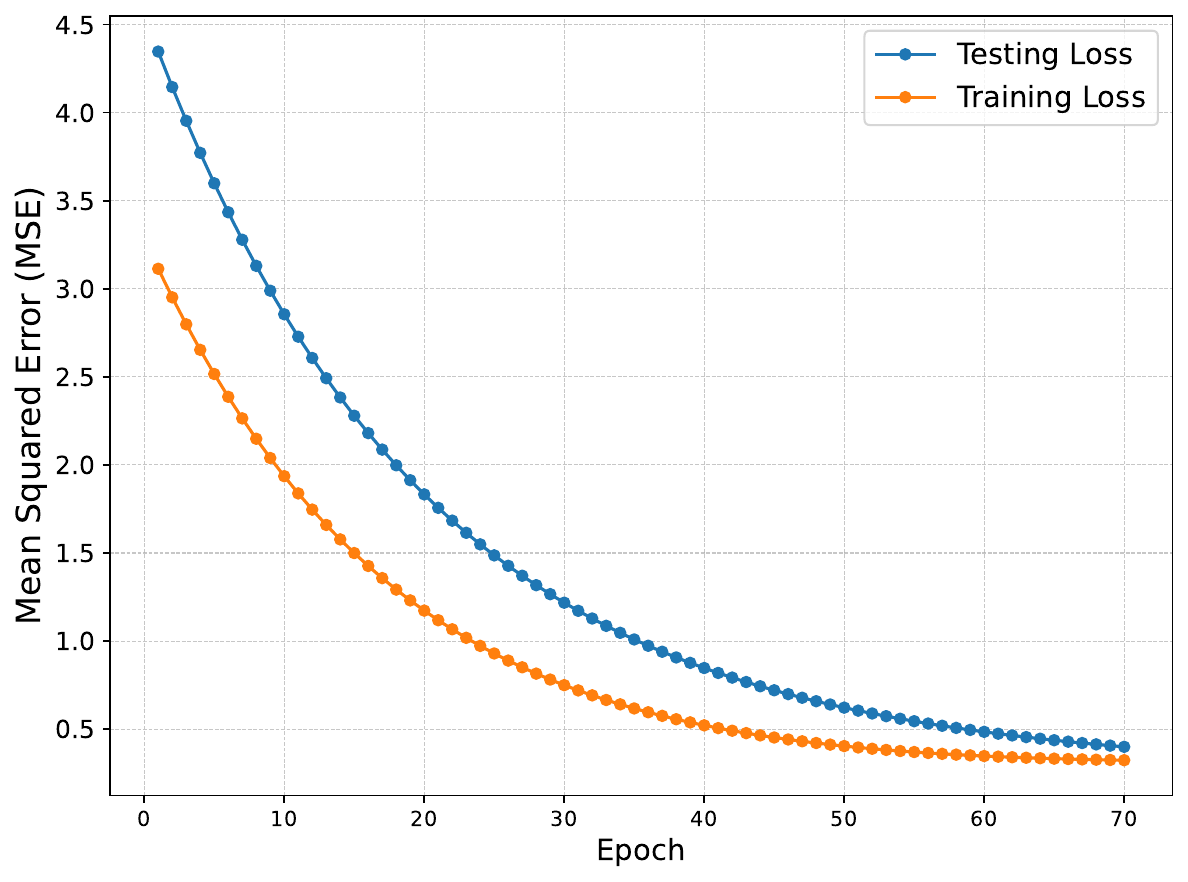}
\end{tabular}
\caption{Training and testing learning curves of the centralized LLaMA model (left) and FedTime (right).}
\label{Fig:Centralized-LLaMA2_versus_FedTime}
\end{figure}

\medskip\noindent\textbf{Federated Performance.}\quad In contrast to the centralized training of the LLaMA model, federated learning adopts a decentralized methodology, allowing for localized model updates and only transmitting the model weights, thus enhancing data privacy. In Figure~\ref{Fig:Centralized-LLaMA2_versus_FedTime}, the training and testing learning curves show that our federated LLM model converges approximately 3$\times$ faster than its centralized counterpart under analogous experimental conditions. We see that while the centralized LLaMA model faltered in convergence even post 200 epochs, the federated FedTime model adeptly minimized the loss function in merely 70 epochs. Also, the federated model's uniform training trajectory, which is largely attributed to the incorporation of a cohesive cluster of edge devices, diverges sharply from the inconsistency seen in the centralized counterpart.

For the sake of federated comparison, we implemented a federated version of PatchTST~\cite{nie2022time}, which we refer to as Fed-PatchTST. We evaluate the performance of FedTime against federated stacked LSTM (FSLSTM)~\cite{abdelsater2021federated} and Fed-PatchTST, and the results are reported in Table~\ref{tab:performance_comparison}, which shows that FedTime consistently outperforms these federated baselines across all the forecasting benchmark datasets. This better performance is particularly notable for longer prediction horizons, highlighting the robustness and adaptability of FedTime to long-term time series forecasting. For instance, the percentage gap and percentage difference between FedTime and Fed-PatchTST, in terms of MAE, on the Electricity dataset are 39.24\% and 32.8\%, respectively, indicating a significant improvement in prediction accuracy with our model.

\begin{table}[!htbp]
\bigskip\bigskip
\caption{Performance comparison of FedTime and federated models across benchmark datasets with prediction length $T=720$.}
\bigskip
\setlength\tabcolsep{4pt} 
\centering
\begin{tabular}{l c c c c c c}
\toprule
Method & \multicolumn{2}{c}{Fed-PatchTST} & \multicolumn{2}{c}{FSLSTM} & \multicolumn{2}{c}{\textbf{FedTime}} \\
\cmidrule(lr){2-3} \cmidrule(lr){4-5} \cmidrule(lr){6-7}
Metric & MSE & MAE & MSE & MAE & MSE & MAE \\
\midrule
Weather & 0.363 & 0.495 & 0.451 & 0.421 & \textbf{0.318} & \textbf{0.335} \\
Traffic & 0.388 & 0.272 & 0.450 & 0.386 & \textbf{0.369} & \textbf{0.239} \\
Electricity & 0.231 & 0.401 & 0.310 & 0.466 & \textbf{0.176} & \textbf{0.288} \\
ETTh1 & 0.456 & 0.520 & 0.522 & 0.621 & \textbf{0.414} & \textbf{0.436} \\
ETTh2 & 0.402 & 0.521 & 0.452 & 0.533 & \textbf{0.376} & \textbf{0.428} \\
ETTm1 & 0.340 & 0.421 & 0.352 & 0.483 & \textbf{0.328} & \textbf{0.373} \\
ETTm2 & 0.335 & 0.384 & 0.401 & 0.495 & \textbf{0.305} & \textbf{0.352} \\
\bottomrule
\end{tabular}
\label{tab:performance_comparison}
\end{table}

\subsection{Communication Overhead}
\begin{figure}[!htb]
\centering
\includegraphics[scale=0.55]{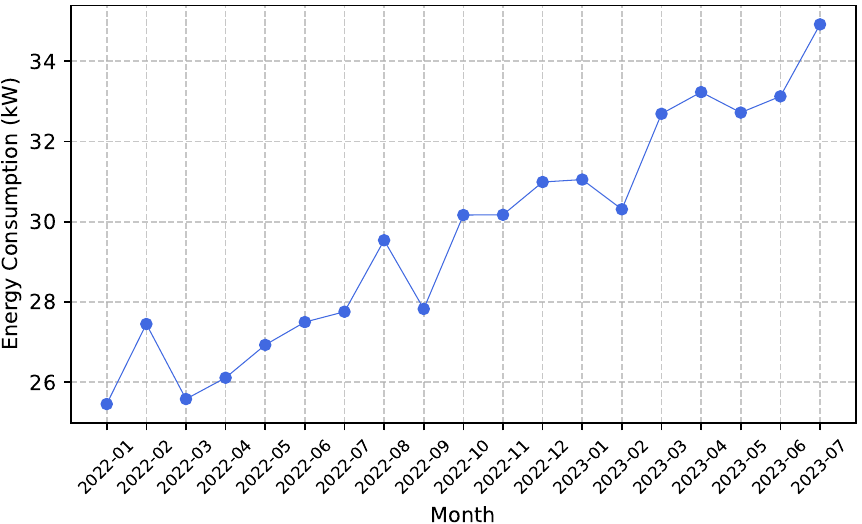}
\caption{ACN energy delivery distribution based on disconnection time.}
\label{Fig:Data_distribution}
\end{figure}

Evaluating the communication overhead is crucial in assessing the feasibility of a model's deployment in federated environments. We use the real-world ACN dataset~\cite{leeacndata2019}, which is an adaptive charging network dataset containing information about an electric vehicle (EV) single charging session at two distinct sites: California Institute of Technology (Caltech 540 stations) and Jet Propulsion Laboratory (JPL 400 stations). An exploratory data analysis of the dataset shows that the data distribution has an increasing trend at both Caltech and JPL sites. These upward trends in the distribution of energy demand at both sites underscores the need for enhanced capacity to accommodate these surges and mitigate escalated demand charges. The ACN dataset consists of 1.5 million charging sessions with a window time of 19 months. Both Caltech and JPL sites show a repetitive usage pattern with much higher utilization during weekdays than on weekends, as shown in Figure~\ref{Fig:Data_distribution}, which depicts the distribution of the energy delivered.

\begin{figure}[!htb]
\centering
\includegraphics[scale=0.36]{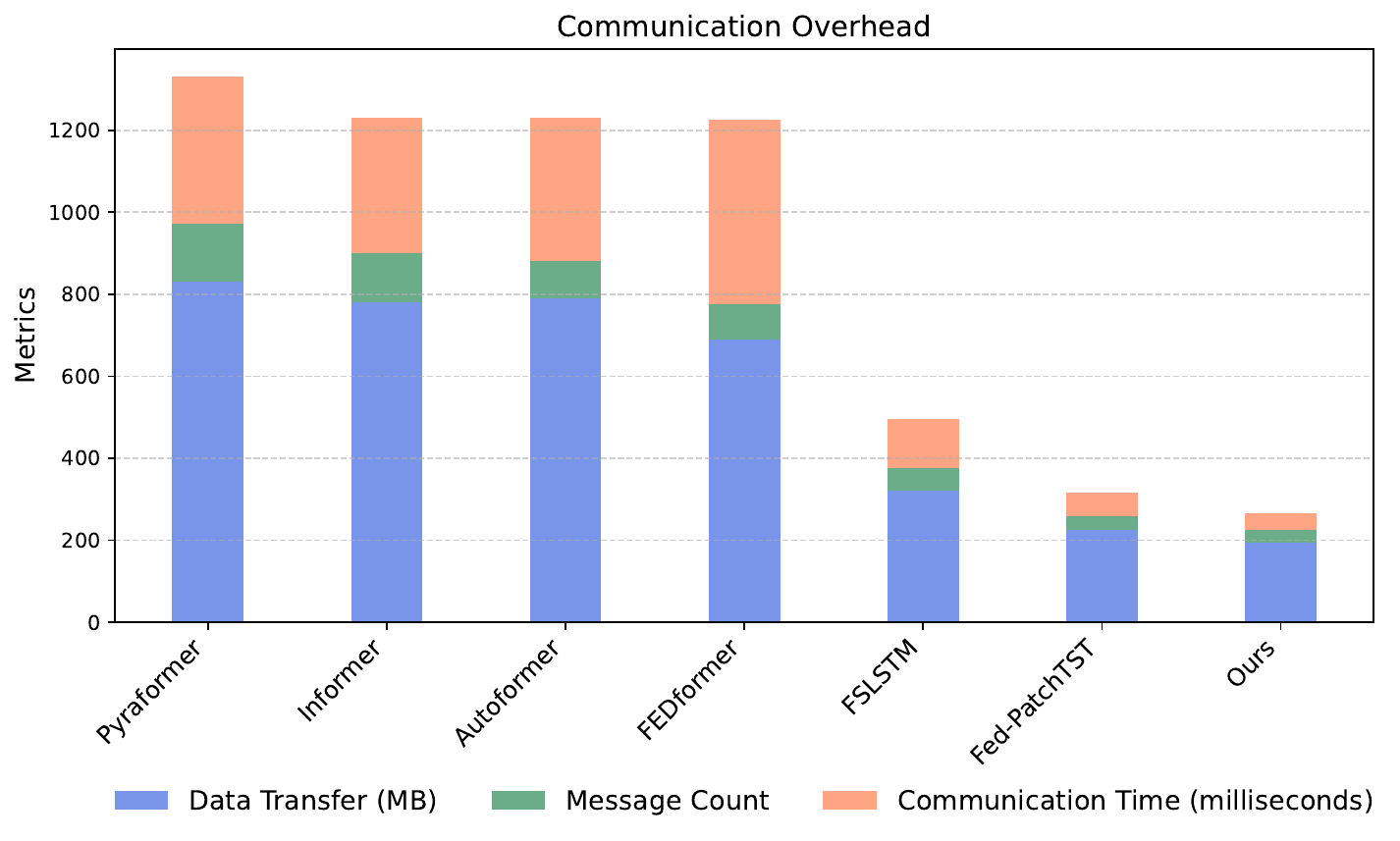}
\caption{Communication overhead comparison between FedTime and baseline methods.}
\label{Fig:Communication_cost}
\end{figure}

One of the key metrics of communication overhead is the data volume transferred, measured in Megabytes. As shown in Figure~\ref{Fig:Communication_cost}, our FedTime model yields reduced communication overhead compared to the baseline methods. This reduction is even more pronounced when considering metrics like message count and communication time. A key factor contributing to FedTime's efficient performance is the strategic engagement of edge devices in the learning process. This decentralized approach, leveraging the inherent distributed nature of IoT-enabled devices, facilitates a system where updating model weights incurs a significantly lower overhead than transferring entire datasets. Such an optimized model structure and design strategy make FedTime notably efficient in comparison with centralized architectures such as Informer, Pyraformer, Autoformer, and LLM4TS. Moreover, Figure~\ref{Fig:Communication_cost} alludes to the efficiency of models with optimized weight structures using PEFT techniques.

\subsection{Ablation Study}\label{Sec:ablation}
\begin{figure}[!htb]
\centering
\includegraphics[scale=0.14]{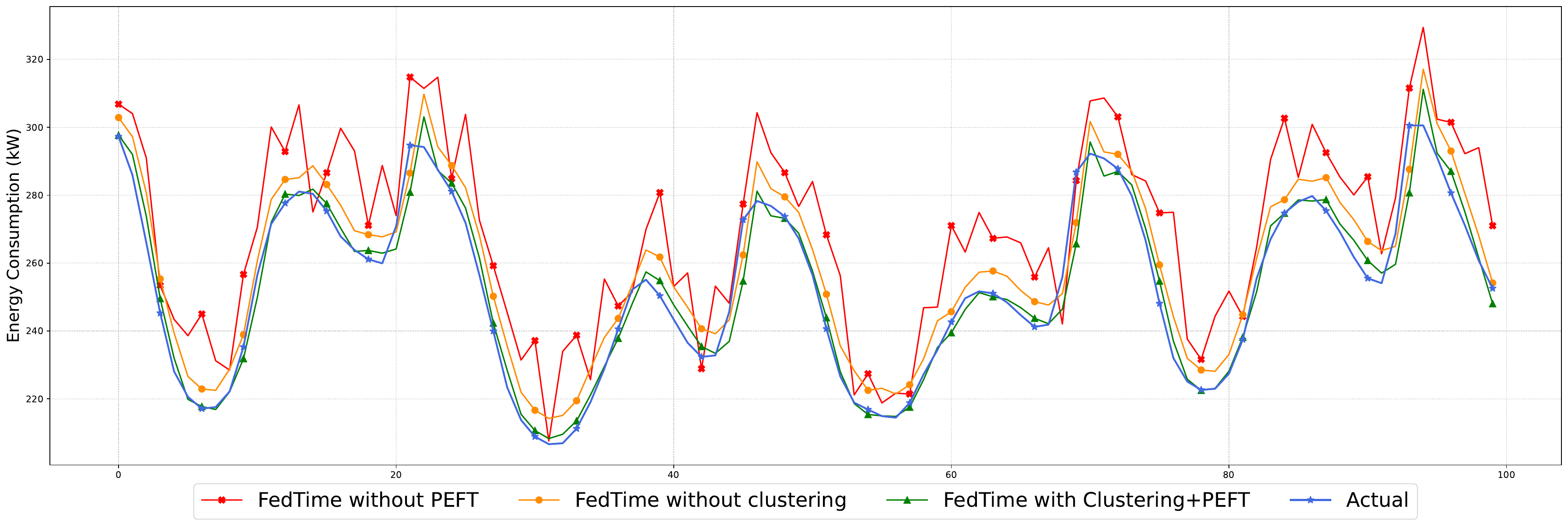}
\caption{Comparison of predicted and actual energy consumption on Caltech site during weekdays (100-hour timestep) using FedTime variants. The horizontal axis represents the different time periods in hours.}
\label{Fig:no_finetuning_vs_full_version}
\end{figure}

In Figure~\ref{Fig:no_finetuning_vs_full_version}, we evaluate the performance of various FedTime variants on the ACN dataset for Caltech site. In the following, we introduce these variants and analyze their respective performances:

\begin{itemize}
\item \textbf{FedTime without Clustering:} This model aligns relatively well with the actual energy consumption in general up to the 20-hour mark. Yet, between the 20- and 40-hour intervals, the model tends to underestimate consumption. Between 60 to 80 hours, the model's predictions come closer to the actual data, but discrepancies are noticeable toward the end of the 100-hour period. The curve's fluctuations are likely due to the mix of devices from different-sized clusters participating in this specific training round.

\item \textbf{FedTime without PEFT:} The performance without parameter efficient fine-tuning generally tracks the consumption trend. Yet, certain intervals display noticeable over-estimations, and there are slight variances from the actual energy consumption, especially during peak and trough periods.

\item \textbf{FedTime with Clustering+PEFT:} This variant manifests a notable performance over the 100-hour duration. Distinctly evident is its aptitude in closely mirroring the authentic consumption, especially in regions characterized by pronounced fluctuations, such as the 20-40 and 60-80 hour intervals. The superior performance of this model makes it the best choice for energy consumption prediction over the scrutinized 100-hour interval at Caltech site.
\end{itemize}

\section{Conclusion}\label{sec:conclusion}
We introduced FedTime, a federated large language model for long-term time series forecasting. FedTime leverages federated learning, ensuring decentralized model training while upholding data privacy across edge devices. Our empirical evaluations underscore FedTime's superiority over both centralized and federated baselines in long-term time series forecasting, demonstrating significant performance enhancements across diverse datasets. Moreover, FedTime addresses communication overhead challenges by optimizing the transmission of model updates, thereby minimizing resource consumption while enhancing forecasting accuracy. In summary, FedTime offers a versatile and efficient solution that strikes a balance between privacy preservation, communication efficiency, and forecasting accuracy in decentralized settings. Looking ahead, our future research endeavors will explore synergies between federated frameworks and other distributed systems, with a keen focus on integrating blockchain technology.


\begin{ack}
This work was supported in part by the Discovery Grants Program of the Natural Sciences and Engineering Research Council of Canada under grant RGPIN-2024-04291.
\end{ack}



\bibliography{references}

\begin{thebibliography}{33}
\providecommand{\natexlab}[1]{#1}
\providecommand{\url}[1]{\texttt{#1}}
\expandafter\ifx\csname urlstyle\endcsname\relax
  \providecommand{\doi}[1]{doi: #1}\else
  \providecommand{\doi}{doi: \begingroup \urlstyle{rm}\Url}\fi

\bibitem[Abdel-Sater and Hamza(2021)]{abdelsater2021federated}
R.~Abdel-Sater and A.~B. Hamza.
\newblock A federated learning approach to anomaly detection in smart
  buildings.
\newblock \emph{ACM Transactions on Internet of Things}, 2\penalty0
  (4):\penalty0 1--23, 2021.

\bibitem[Challu et~al.(2023)Challu, Olivares, Oreshkin, Ramirez, Canseco, and
  Dubrawski]{challu2023nhits}
C.~Challu, K.~G. Olivares, B.~N. Oreshkin, F.~G. Ramirez, M.~M. Canseco, and
  A.~Dubrawski.
\newblock {N-HiTS}: Neural hierarchical interpolation for time series
  forecasting.
\newblock In \emph{Proc. AAAI Conference on Artificial Intelligence}, pages
  6989--6997, 2023.

\bibitem[Chang et~al.(2023)Chang, Peng, and Chen]{Chang2023LLM4TS}
C.~Chang, W.-C. Peng, and T.-F. Chen.
\newblock {LLM4TS}: Two-stage fine-tuning for time-series forecasting with
  pre-trained {LLMs}.
\newblock \emph{arXiv preprint arXiv:2308.08469}, 2023.

\bibitem[Chen et~al.(2021)Chen, Peng, Fu, and Ling]{chen2021autoformer}
M.~Chen, H.~Peng, J.~Fu, and H.~Ling.
\newblock Autoformer: Searching transformers for visual recognition.
\newblock In \emph{Proc. IEEE International Conference on Computer Vision},
  pages 12270--12280, 2021.

\bibitem[Cui et~al.(2023)Cui, Yuan, Ding, Yao, Zhu, Ni, Xie, Liu, and
  Sun]{cui2023ultrafeedback}
G.~Cui, L.~Yuan, N.~Ding, G.~Yao, W.~Zhu, Y.~Ni, G.~Xie, Z.~Liu, and M.~Sun.
\newblock {UltraFeedback}: Boosting language models with high-quality feedback.
\newblock In \emph{arXiv preprint arXiv:2310.01377}, 2023.

\bibitem[Das et~al.(2023)Das, Kong, Leach, Sen, and Yu]{das2023long}
A.~Das, W.~Kong, A.~Leach, R.~Sen, and R.~Yu.
\newblock Long-term forecasting with {TiDE}: Time-series dense encoder.
\newblock \emph{Transactions on Machine Learning Research}, 2023.

\bibitem[Dettmers et~al.(2023)Dettmers, Pagnoni, Holtzman, and
  Zettlemoyer]{dettmers2023qlora}
T.~Dettmers, A.~Pagnoni, A.~Holtzman, and L.~Zettlemoyer.
\newblock {QLoRA}: Efficient finetuning of quantized {LLM}s.
\newblock In \emph{Advances in Neural Information Processing Systems}, 2023.

\bibitem[Hu et~al.(2022)Hu, Shen, Wallis, Allen-Zhu, Li, Wang, Wang, and
  Chen]{hu2021lora}
E.~J. Hu, Y.~Shen, P.~Wallis, Z.~Allen-Zhu, Y.~Li, S.~Wang, L.~Wang, and
  W.~Chen.
\newblock {LoRA}: Low-rank adaptation of large language models.
\newblock In \emph{International Conference on Learning Representations}, 2022.

\bibitem[Kairouz et~al.()Kairouz, McMahan, Avent, Bellet, Bennis, Bhagoji,
  Bonawitz, Charles, Cormode, Cummings, et~al.]{kairouz2021advances}
P.~Kairouz, H.~B. McMahan, B.~Avent, A.~Bellet, M.~Bennis, A.~N. Bhagoji,
  K.~Bonawitz, Z.~Charles, G.~Cormode, R.~Cummings, et~al.
\newblock Advances and open problems in federated learning.
\newblock \emph{Foundations and Trends{\textregistered} in Machine Learning}.

\bibitem[Kim et~al.(2022)Kim, Kim, Tae, Park, Choi, and Choo]{Kim2022RevIN}
T.~Kim, J.~Kim, Y.~Tae, C.~Park, J.-H. Choi, and J.~Choo.
\newblock Reversible instance normalization for accurate time-series
  forecasting against distribution shift.
\newblock In \emph{International Conference on Learning Representations}, 2022.

\bibitem[Lee et~al.(2019)Lee, Li, and Low]{leeacndata2019}
Z.~J. Lee, T.~Li, and S.~H. Low.
\newblock {ACN-Data: Analysis and applications of an open {EV} charging
  dataset}.
\newblock In \emph{Proc. International Conference on Future Energy Systems},
  2019.

\bibitem[Li et~al.(2021)Li, Wen, Wu, Hu, Wang, Li, Liu, and He]{li2021survey}
Q.~Li, Z.~Wen, Z.~Wu, S.~Hu, N.~Wang, Y.~Li, X.~Liu, and B.~He.
\newblock A survey on federated learning systems: Vision, hype and reality for
  data privacy and protection.
\newblock \emph{IEEE Transactions on Knowledge and Data Engineering}, 2021.

\bibitem[Li et~al.(2022)Li, Diao, Chen, and He]{li2022federated}
Q.~Li, Y.~Diao, Q.~Chen, and B.~He.
\newblock Federated learning on non-{IID} data silos: An experimental study.
\newblock In \emph{Proc. IEEE International Conference on Data Engineering},
  pages 965--978, 2022.

\bibitem[Liu et~al.(2021)Liu, Yu, Liao, Li, Lin, Liu, and
  Dustdar]{liu2021pyraformer}
S.~Liu, H.~Yu, C.~Liao, J.~Li, W.~Lin, A.~X. Liu, and S.~Dustdar.
\newblock Pyraformer: Low-complexity pyramidal attention for long-range time
  series modeling and forecasting.
\newblock In \emph{International Conference on Learning Representations}, 2021.

\bibitem[McMahan et~al.(2017)McMahan, Moore, Ramage, Hampson, and
  y~Arcas]{mcmahan2017communication}
B.~McMahan, E.~Moore, D.~Ramage, S.~Hampson, and B.~A. y~Arcas.
\newblock Communication-efficient learning of deep networks from decentralized
  data.
\newblock In \emph{Proc. International Conference on Artificial Intelligence
  and Statistics}, pages 1273--1282, 2017.

\bibitem[Mills et~al.(2019)Mills, Hu, and Min]{mills2019communication}
J.~Mills, J.~Hu, and G.~Min.
\newblock Communication-efficient federated learning for wireless edge
  intelligence in {I}o{T}.
\newblock \emph{IEEE Internet of Things Journal}, 7\penalty0 (7):\penalty0
  5986--5994, 2019.

\bibitem[Murugesan and Cherukuri(2023)]{murugesan2023rise}
S.~Murugesan and A.~K. Cherukuri.
\newblock The rise of generative artificial intelligence and its impact on
  education: The promises and perils.
\newblock \emph{Computer}, 56\penalty0 (5):\penalty0 116--121, 2023.

\bibitem[Nie et~al.(2023)Nie, Nguyen, Sinthong, and Kalagnanam]{nie2022time}
Y.~Nie, N.~H. Nguyen, P.~Sinthong, and J.~Kalagnanam.
\newblock A time series is worth 64 words: long-term forecasting with
  {T}ransformers.
\newblock In \emph{International Conference on Learning Representations}, 2023.

\bibitem[Ouyang et~al.(2022)Ouyang, Wu, Jiang, Almeida, Wainwright, Mishkin,
  Zhang, Agarwal, Slama, Ray, et~al.]{ouyang2022training}
L.~Ouyang, J.~Wu, X.~Jiang, D.~Almeida, C.~Wainwright, P.~Mishkin, C.~Zhang,
  S.~Agarwal, K.~Slama, A.~Ray, et~al.
\newblock Training language models to follow instructions with human feedback.
\newblock In \emph{Advances in Neural Information Processing Systems}, pages
  27730--27744, 2022.

\bibitem[Rafailov et~al.(2023)Rafailov, Sharma, Mitchell, Ermon, Manning, and
  Finn]{rafailov2023direct}
R.~Rafailov, A.~Sharma, E.~Mitchell, S.~Ermon, C.~D. Manning, and C.~Finn.
\newblock Direct preference optimization: Your language model is secretly a
  reward model.
\newblock In \emph{Advances in Neural Information Processing Systems}, 2023.

\bibitem[Reddi et~al.(2021)Reddi, Charles, Zaheer, Garrett, Rush,
  Kone{\v{c}}n{\'y}, Kumar, and McMahan]{reddi2021fedAdam}
S.~J. Reddi, Z.~Charles, M.~Zaheer, Z.~Garrett, K.~Rush, J.~Kone{\v{c}}n{\'y},
  S.~Kumar, and H.~B. McMahan.
\newblock Adaptive federated optimization.
\newblock In \emph{International Conference on Learning Representations}, 2021.

\bibitem[Saputra et~al.(2019)Saputra, Hoang, Nguyen, Dutkiewicz, Mueck, and
  Srikanteswara]{Saputra2019FEDL}
Y.~M. Saputra, D.~T. Hoang, D.~N. Nguyen, E.~Dutkiewicz, M.~D. Mueck, and
  S.~Srikanteswara.
\newblock Energy demand prediction with federated learning for electric vehicle
  networks.
\newblock In \emph{Proc. IEEE Global Communications Conference}, 2019.

\bibitem[Shazeer(2020)]{shazeer2020glu}
N.~Shazeer.
\newblock {GLU} variants improve {T}ransformer.
\newblock \emph{arXiv preprint arXiv:2002.05202}, 2020.

\bibitem[Taori et~al.(2023)Taori, Gulrajani, Zhang, Dubois, Li, Guestrin,
  Liang, and Hashimoto]{taori2023stanford}
R.~Taori, I.~Gulrajani, T.~Zhang, Y.~Dubois, X.~Li, C.~Guestrin, P.~Liang, and
  T.~B. Hashimoto.
\newblock {Stanford Alpaca}: An instruction-following {LLaMA} model, 2023.

\bibitem[Thoppilan et~al.(2022)Thoppilan, De~Freitas, Hall, Shazeer,
  Kulshreshtha, Cheng, Jin, Bos, Baker, Du, et~al.]{thoppilan2022lamda}
R.~Thoppilan, D.~De~Freitas, J.~Hall, N.~Shazeer, A.~Kulshreshtha, H.-T. Cheng,
  A.~Jin, T.~Bos, L.~Baker, Y.~Du, et~al.
\newblock {LaMDA}: Language models for dialog applications.
\newblock \emph{arXiv preprint arXiv:2201.08239}, 2022.

\bibitem[Touvron et~al.(2023)Touvron, Lavril, Izacard, Martinet, Lachaux,
  Lacroix, Rozi{\`e}re, Goyal, Hambro, Azhar, et~al.]{touvron2023llama}
H.~Touvron, T.~Lavril, G.~Izacard, X.~Martinet, M.-A. Lachaux, T.~Lacroix,
  B.~Rozi{\`e}re, N.~Goyal, E.~Hambro, F.~Azhar, et~al.
\newblock {LLaMA}: Open and efficient foundation language models.
\newblock \emph{arXiv preprint arXiv:2302.13971}, 2023.

\bibitem[Wen et~al.(2023)Wen, Zhang, Lan, Cui, Cai, and Zhang]{wen2023survey}
J.~Wen, Z.~Zhang, Y.~Lan, Z.~Cui, J.~Cai, and W.~Zhang.
\newblock A survey on federated learning: challenges and applications.
\newblock \emph{International Journal of Machine Learning and Cybernetics},
  14\penalty0 (2):\penalty0 513--535, 2023.

\bibitem[Wu et~al.(2021)Wu, Xu, Wang, and Long]{wu2021autoformer}
H.~Wu, J.~Xu, J.~Wang, and M.~Long.
\newblock Autoformer: Decomposition {T}ransformers with auto-correlation for
  long-term series forecasting.
\newblock In \emph{Advances in Neural Information Processing Systems}, pages
  22419--22430, 2021.

\bibitem[Zeng et~al.(2023)Zeng, Chen, Zhang, and Xu]{zeng2023transformers}
A.~Zeng, M.~Chen, L.~Zhang, and Q.~Xu.
\newblock Are {T}ransformers effective for time series forecasting?
\newblock In \emph{Proc. AAAI conference on Artificial Intelligence}, pages
  11121--11128, 2023.

\bibitem[Zhang et~al.(2023)Zhang, Zhang, Li, Zhao, Karypis, and
  Smola]{zhang2023multimodal}
Z.~Zhang, A.~Zhang, M.~Li, H.~Zhao, G.~Karypis, and A.~Smola.
\newblock Multimodal chain-of-thought reasoning in language models.
\newblock \emph{arXiv preprint arXiv:2302.00923}, 2023.

\bibitem[Zheng et~al.(2023)Zheng, Chiang, Sheng, Zhuang, Wu, Zhuang, Lin, Li,
  Li, Xing, et~al.]{zheng2023judging}
L.~Zheng, W.-L. Chiang, Y.~Sheng, S.~Zhuang, Z.~Wu, Y.~Zhuang, Z.~Lin, Z.~Li,
  D.~Li, E.~Xing, et~al.
\newblock Judging {LLM}-as-a-judge with {MT}-bench and chatbot arena.
\newblock In \emph{Advances in Neural Information Processing Systems}, 2023.

\bibitem[Zhou et~al.(2021)Zhou, Zhang, Peng, Zhang, Li, Xiong, and
  Zhang]{zhou2021informer}
H.~Zhou, S.~Zhang, J.~Peng, S.~Zhang, J.~Li, H.~Xiong, and W.~Zhang.
\newblock Informer: Beyond efficient {T}ransformer for long sequence
  time-series forecasting.
\newblock In \emph{Proc. AAAI Conference on Artificial Intelligence}, pages
  11106--11115, 2021.

\bibitem[Zhou et~al.(2022)Zhou, Ma, Wen, Wang, Sun, and Jin]{zhou2022fedformer}
T.~Zhou, Z.~Ma, Q.~Wen, X.~Wang, L.~Sun, and R.~Jin.
\newblock {FEDformer}: Frequency enhanced decomposed {T}ransformer for
  long-term series forecasting.
\newblock In \emph{Proc. International Conference on Machine Learning}, pages
  27268--27286, 2022.

\end{thebibliography}

\end{document}